\pdfoutput=1

\documentclass[11pt]{article}

\usepackage{ACL2023}
\usepackage{times}
\usepackage{latexsym}

\usepackage[T1]{fontenc}

\usepackage[utf8]{inputenc}

\usepackage{microtype}

\usepackage{inconsolata}
\usepackage{url}

\usepackage{amsmath}
\usepackage{multirow}
\usepackage{xcolor}

\usepackage{xcolor,colortbl}
\definecolor{lightblue}{rgb}{0.68, 0.85, 0.9}
\definecolor{lavender}{rgb}{0.9, 0.9, 0.98}
\definecolor{lightyellow}{rgb}{1.0, 1.0, 0.88}
\definecolor{magicmint}{rgb}{0.67, 0.94, 0.82}
\definecolor{palepink}{rgb}{0.98, 0.85, 0.87}
\definecolor{bubbles}{rgb}{0.91, 1.0, 1.0}

\usepackage{microtype}
\usepackage{graphicx}
\usepackage{subfigure}
\usepackage{amssymb}
\usepackage{CJKutf8}
\usepackage{algorithm}
\usepackage{algorithmic}
\usepackage{enumitem}
\algsetup{linenosize=\small}
\usepackage{wrapfig}
\usepackage{cleveref}
\usepackage{multicol}
\usepackage{booktabs}
\usepackage{xspace}
\usepackage{adjustbox}
\usepackage{subfiles}
\usepackage{latexsym}

\usepackage{placeins}
\usepackage{tikz}

\newcommand*{\affaddr}[1]{#1} 
\newcommand*{\affmark}[1][*]{\textsuperscript{#1}}
\newcommand*{\email}[1]{\textrm{#1}}

\usepackage{graphicx} 

\title{Can ChatGPT-like Generative Models Guarantee Factual Accuracy? \\On the Mistakes of New Generation Search Engines}

\author{Ruochen Zhao\thanks{\; Equal contribution, order decided by coin flip.
 }\affmark[~~1]~~Xingxuan Li\footnotemark[1]\affmark[~~1]~~Yew Ken Chia\footnotemark[1]~~~~\textbf{Bosheng Ding\footnotemark[1]\affmark[~~1]\affmark[~~]}~~\textbf{Lidong Bing\affmark[2]}~~\\
\affaddr{\affmark[1]Nanyang Technological University, Singapore}
\affaddr{\affmark[2]DAMO Academy, Alibaba Group}\\
\email{\small{\{ruochen002, xingxuan001, bosheng001\}@e.ntu.edu.sg}}\\
\email{\small{chiayewken@gmail.com}} \\
\email{\small{l.bing@alibaba-inc.com}}
}

\begin{document}

\maketitle

\begin{abstract}

Although large conversational AI models such as OpenAI's ChatGPT have demonstrated great potential, we question whether such models can guarantee factual accuracy.
Recently, technology companies such as Microsoft and Google have announced new services which aim to combine search engines with conversational AI.
However, we have found numerous mistakes in the public demonstrations that suggest we should not easily trust the factual claims of the AI models.
Rather than criticizing specific models or companies, we hope to call on researchers and developers to improve AI models'
transparency and factual correctness. \footnote{Our first released blog post can be found \href{https://dev.to/ruochenzhao3/can-chatgpt-like-generative-models-guarantee-factual-accuracy-on-the-mistakes-of-microsofts-new-bing-111b}{here}.}
\end{abstract}

\section{Introduction}

Recently, conversational AI models such as OpenAI's \href{https://openai.com/blog/chatgpt/}{ChatGPT} \citep{chatgptblog} have captured public imagination with the ability to generate high-quality written contents, hold human-like conversations, answer factual questions, and more. Armed with such potential, Microsoft and Google have announced new services \citep{vincent_2023} that combine them with traditional search engines. The new wave of conversation-powered search engines has the potential to naturally answer complex questions, summarize search results, and even serve as a creative tool. However, in doing so, the tech companies now face a greater ethical challenge to ensure that their models do not mislead users with false, ungrounded, or conflicting answers. Hence, the question naturally arises: Can ChatGPT-like models guarantee factual accuracy? In this article, we uncover several factual mistakes in Microsoft's \href{https://www.bing.com/new}{new Bing} \citep{bing} and Google's \href{https://blog.google/technology/ai/bard-google-ai-search-updates/}{Bard} \citep{pichai_2023} which suggest that they currently cannot.

Unfortunately, false expectations can lead to disastrous results. Around the same time as Microsoft's new Bing announcement, Google hastily announced a new conversational AI service named Bard. Despite the hype, expectations were quickly shattered when Bard made a factual mistake in the \href{https://twitter.com/sundarpichai/status/1622673775182626818}{promotional video}, eventually tanking Google's share price \citep{sherman_2023} by nearly 8\% and wiping \$100 billion off its market value. On the other hand, there has been less scrutiny regarding Microsoft's new Bing. In the \href{https://www.youtube.com/watch?v=rOeRWRJ16yY}{demonstration video}, we found that the new Bing recommended a rock singer as a top poet, fabricated birth and death dates, and even made up an entire summary of fiscal reports. Despite disclaimers that the new Bing's responses may not always be factual, overly optimistic sentiments may inevitably lead to disillusionment. Hence, our goal is to draw attention to the factual challenges faced by conversation-powered search engines so that we may better address them in the future.

\section{What Factual Mistakes did Microsoft's New Bing Demonstrate?}

Microsoft released the new Bing search engine powered by AI, claiming that it will revolutionize the scope of traditional search engines. Is this really the case? We dived deeper into the \href{https://www.youtube.com/watch?v=rOeRWRJ16yY}{demonstration video} and \href{https://www.bing.com/new}{examples}, and found three main types of factual issues:
\begin{itemize}[leftmargin=*,topsep=2pt,itemsep=2pt,parsep=0pt]
    \item Claims that conflict with the reference sources.
    \item Claims that don't exist in the reference sources.
    \item Claims that don't have a reference source, and are inconsistent with multiple web sources.
\end{itemize}

\begin{figure}[ht]
\centering
\includegraphics[width=1.0\columnwidth, trim={30cm 0 0 0}, clip]{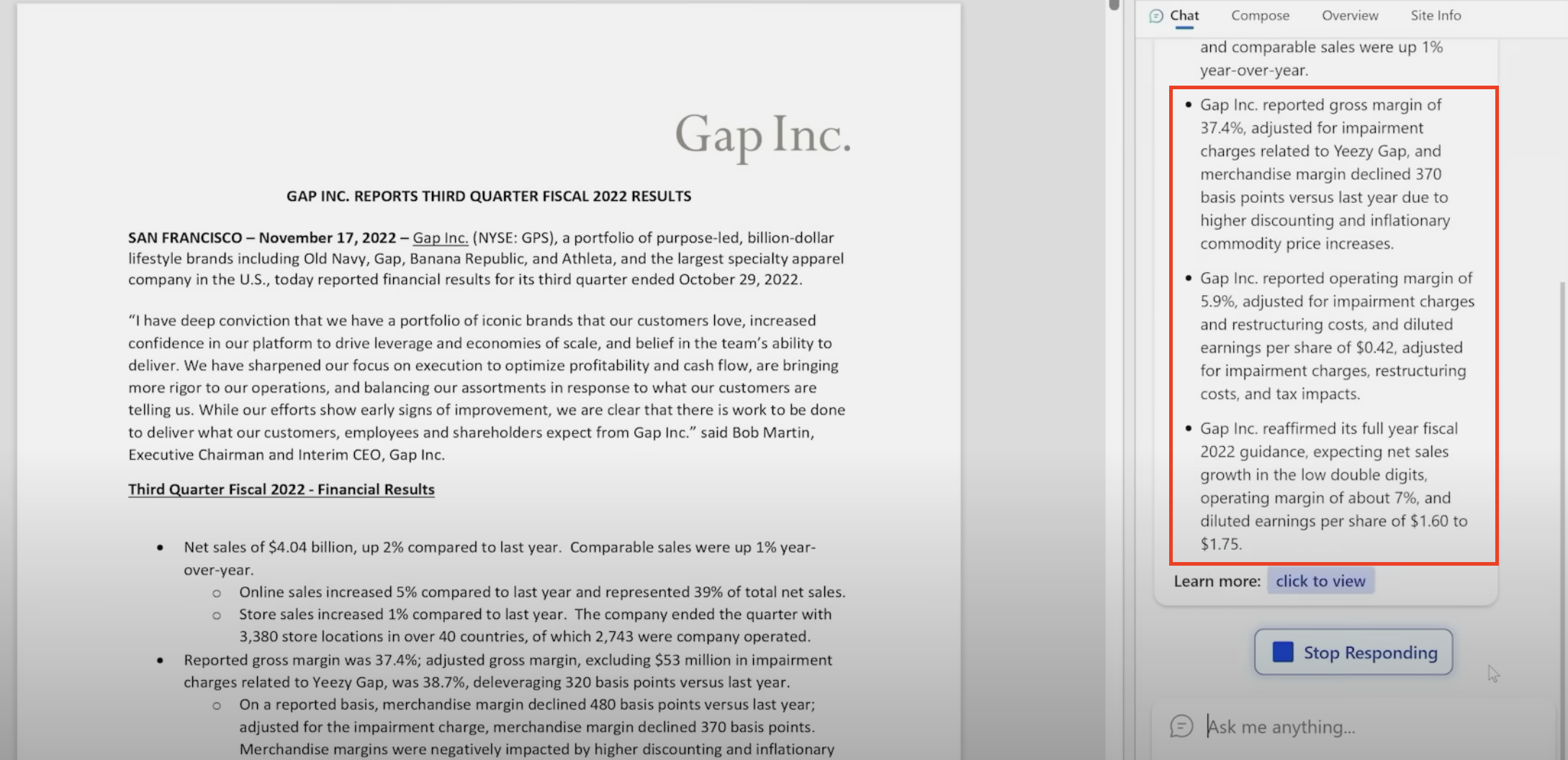}
\caption{Summary of the Gap Inc. fiscal report by the new Bing in press release.}
\label{fig:gap-report}
\end{figure}

\paragraph{Fabricated numbers in the summary of financial reports: be careful when you trust the new Bing!}
To our surprise, the new Bing fabricated an entire summary of the financial report in the demonstration! 
When Microsoft executive Yusuf Mehdi showed the audience how to use the command "key takeaways from the page" to auto-generate a summary of the Gap Inc. 2022 Q3 Fiscal Report \citep{gabreport}, he received the following results as shown in Figure \ref{fig:gap-report}.

\begin{figure}[ht]
\centering
\includegraphics[width=1.0\columnwidth]{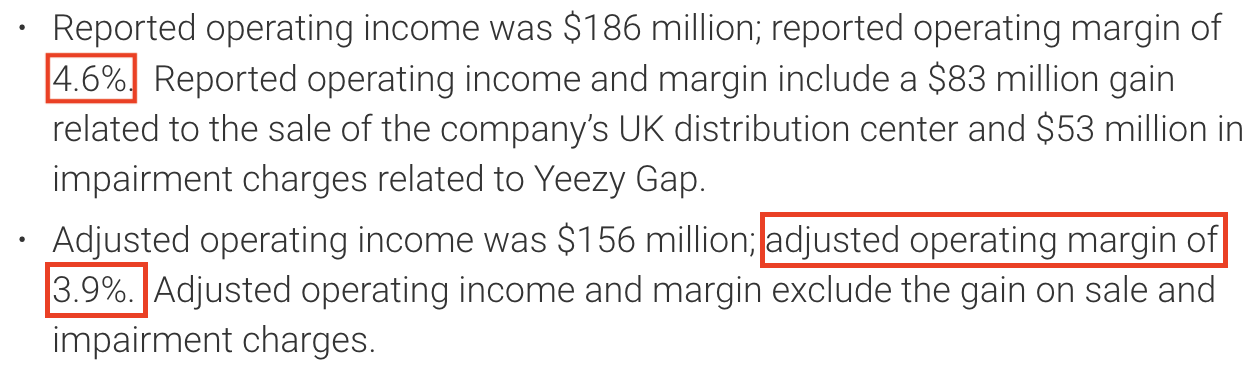}
\caption{Gap Inc. fiscal report excerpt on operating margins.}
\label{fig:gap-margin}
\end{figure}

However, upon closer examination, all the key figures in the generated summary are inaccurate. We will show excerpts from the original financial report below as validating references. According to the new Bing, the operating margin after adjustment was 5.9\%, while it was actually 3.9\% in the source report as shown in Figure \ref{fig:gap-margin}.

\begin{figure}[ht]
\centering
\includegraphics[width=1.0\columnwidth]{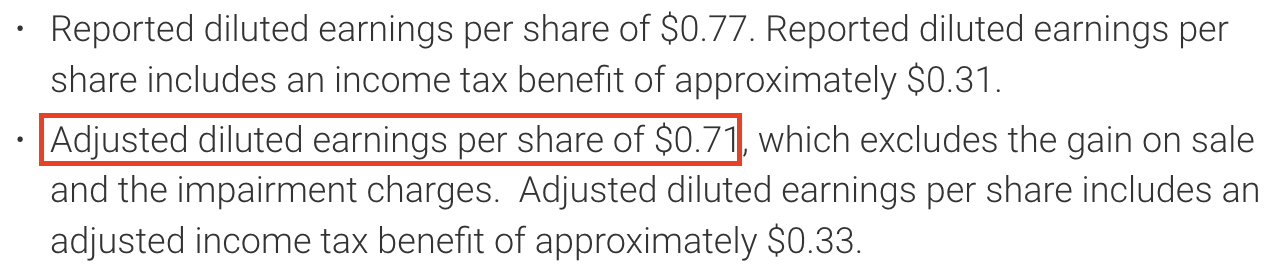}
\caption{Gap Inc. fiscal report excerpt on operating margins.}
\label{fig:gap-earnings}
\end{figure}

Similarly, the adjusted diluted earnings per share was generated as \$0.42, while it should be \$0.71 as shown in Figure \ref{fig:gap-earnings}.

\begin{figure}[ht]
\centering
\includegraphics[width=1.0\columnwidth]{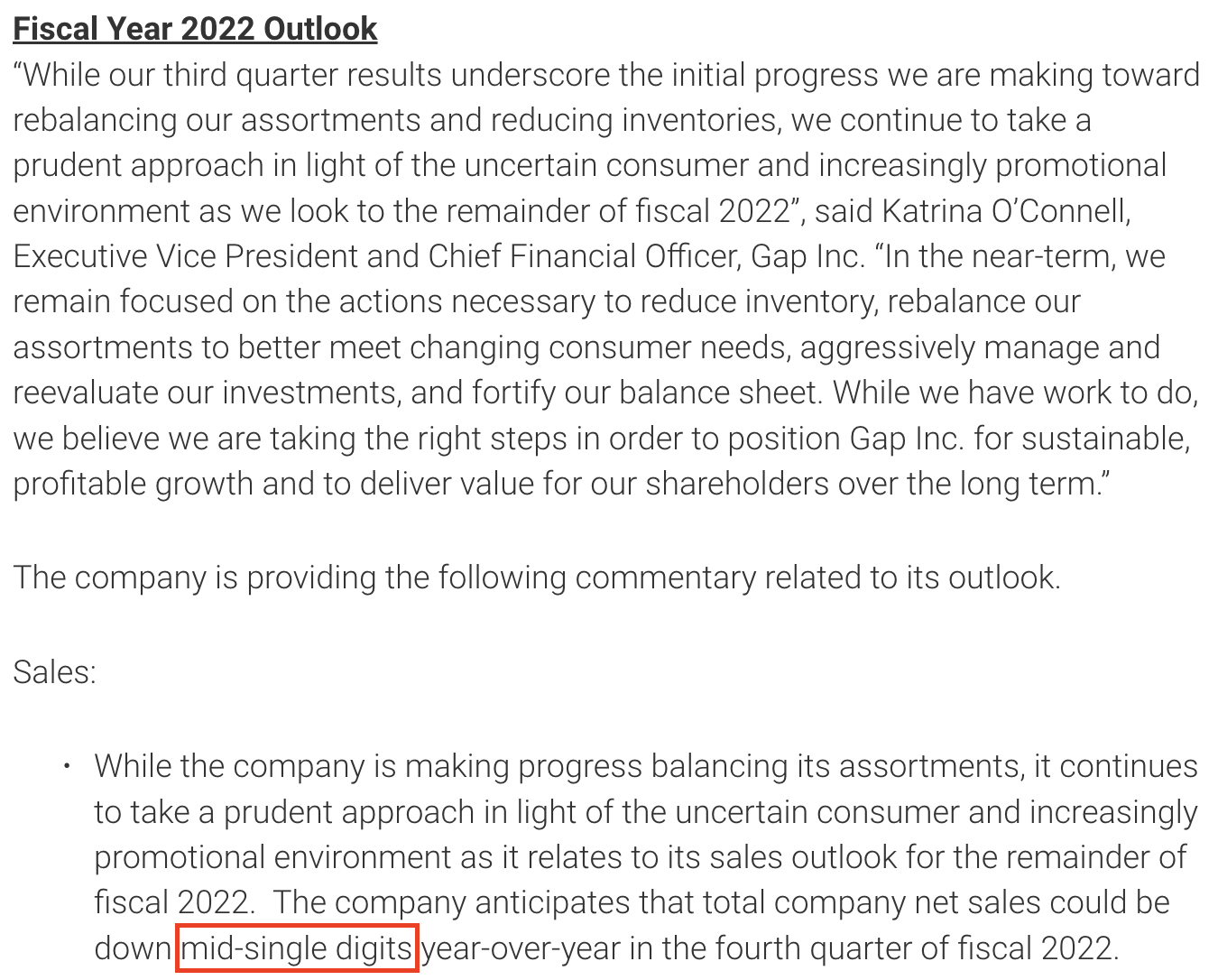}
\caption{Gap Inc. fiscal report on 2022 outlook.}
\label{fig:gap-sales}
\end{figure}

Regarding net sales, the new Bing's summary claimed "growth in the low double digits", while the original report shown in Figure \ref{fig:gap-sales} stated that "net sales could be down mid-single digits".

In addition to the generated figures which conflicted with actual figures in the source report, we observe that the new Bing may also produce hallucinated facts that do not exist in the source. In the new Bing's generated summary, the "operating margin of about 7\% and diluted earnings per share of \$1.60 to \$1.75" are nowhere to be found in the source report.

Unfortunately, the situation worsened when the new Bing was instructed to "compare this with Lululemon in a table". The financial comparison table generated by the new Bing contained numerous mistakes:

\begin{figure}[ht]
\centering
\includegraphics[width=1.0\columnwidth, trim={30cm 0 0 0}, clip]{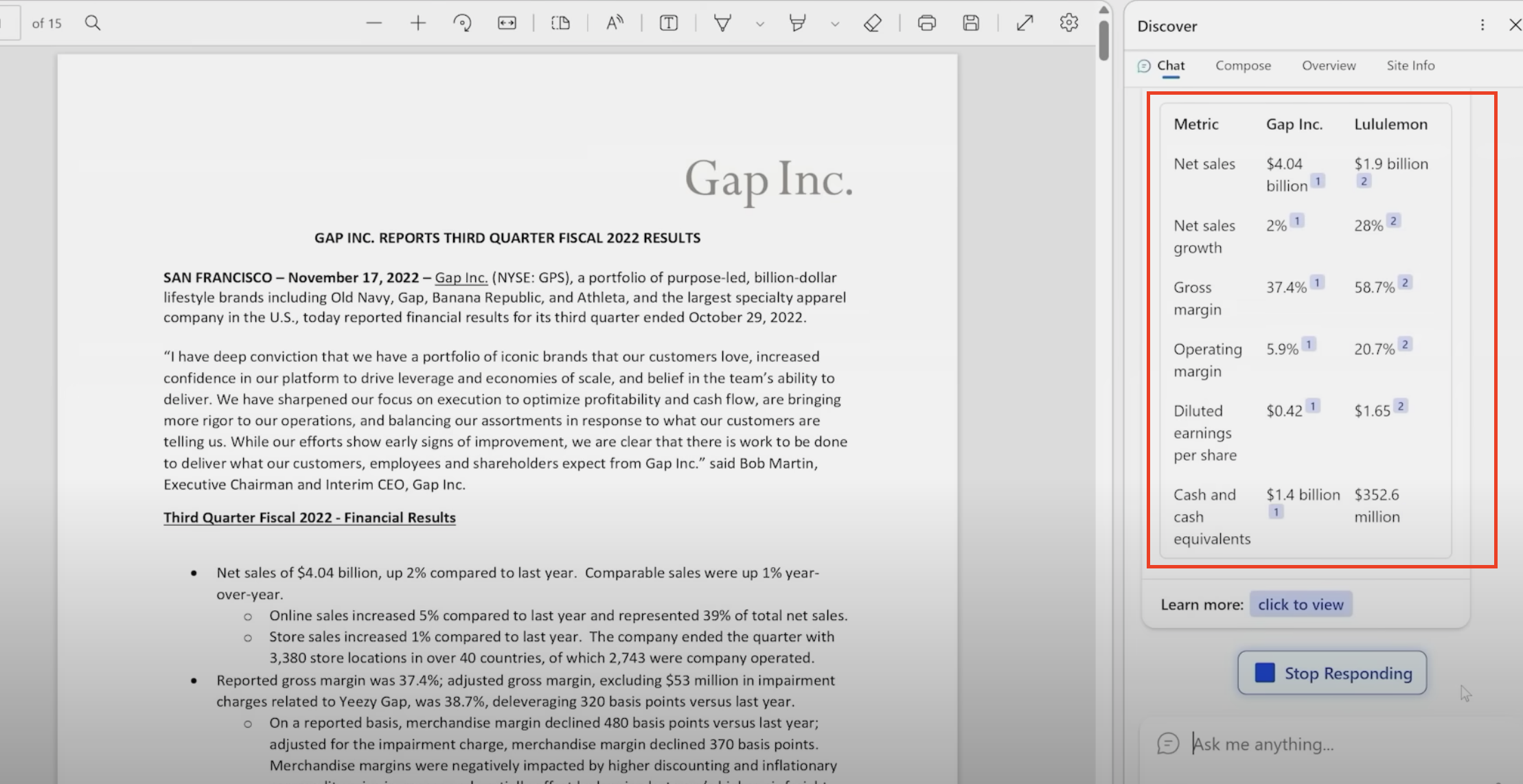}
\caption{The comparison table generated by the new Bing in press release.}
\label{fig:gap-compare}
\end{figure}

\begin{figure}[ht]
\centering
\includegraphics[width=1.0\columnwidth]{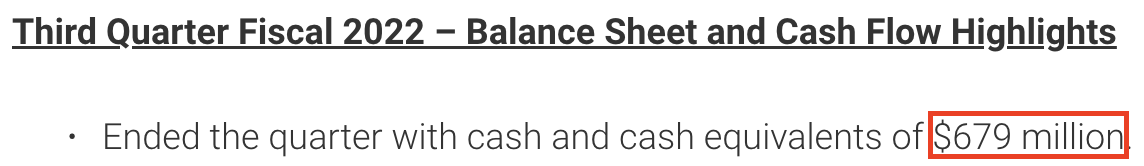}
\caption{Gap Inc. fiscal report excerpt on cash.}
\label{fig:gap-cash}
\end{figure}

This table in Figure \ref{fig:gap-compare}, in fact, is half wrong. Out of all the numbers, 3 out of 6 figures are wrong in the column for Gap Inc., and same for Lululemon. As mentioned before, Gap Inc.'s true operating margin is 4.6\% (or 3.9\% after adjusting) and diluted earnings per share should be \$0.77 (or \$0.71 after adjusting). The new Bing also claimed that Gap Inc.'s cash and cash equivalents amounted to \$1.4 billion, while it was actually \$679 million as shown in Figure \ref{fig:gap-cash}.

\begin{figure}[ht]
\centering
\includegraphics[width=1.0\columnwidth, trim={0 0 15cm 0}, clip]{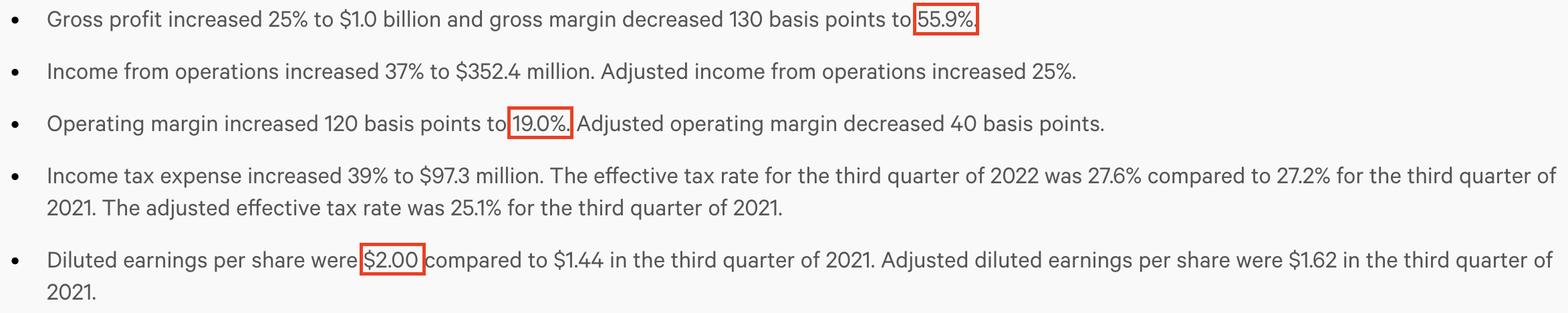}
\caption{Lululemon 2022 Q3 fiscal report excerpt.}
\label{fig:lululemon-report}
\end{figure}

According to Lululemon's 2022 Q3 Fiscal Report \citep{lululemonreport} as shown in Figure \ref{fig:lululemon-report}, the gross margin should be 55.9\%, while the new Bing claims it's 58.7\%. The operating margin should be 19.0\%, while the new Bing claims it to be 20.7\%. The diluted earnings per share was actually \$2.00, while the new Bing claims it to be \$1.65.

So where did these figures come from? You may be wondering whether it's a number that was misplaced from another part in the original document. The answer is no. Curiously, these numbers are nowhere to be found in the original document and are entirely fabricated. In fact, it is still an open research challenge to constrain the outputs of generative models to be more factually grounded. Plainly speaking, the popular generative AI models such as ChatGPT are picking words to generate from a fixed vocabulary, instead of strictly copying and pasting facts from the source. Hence, factual correctness is one of the innate challenges of generative AI, and cannot be strictly guaranteed with current models. This is a major concern when it comes to search engines as users rely on the results to be trustworthy and factually accurate.

\paragraph{Top Japanese poet: secretly a rock singer?}

\begin{figure}[ht]
\centering
\includegraphics[width=1.0\columnwidth, trim={0 0 15cm 0}, clip]{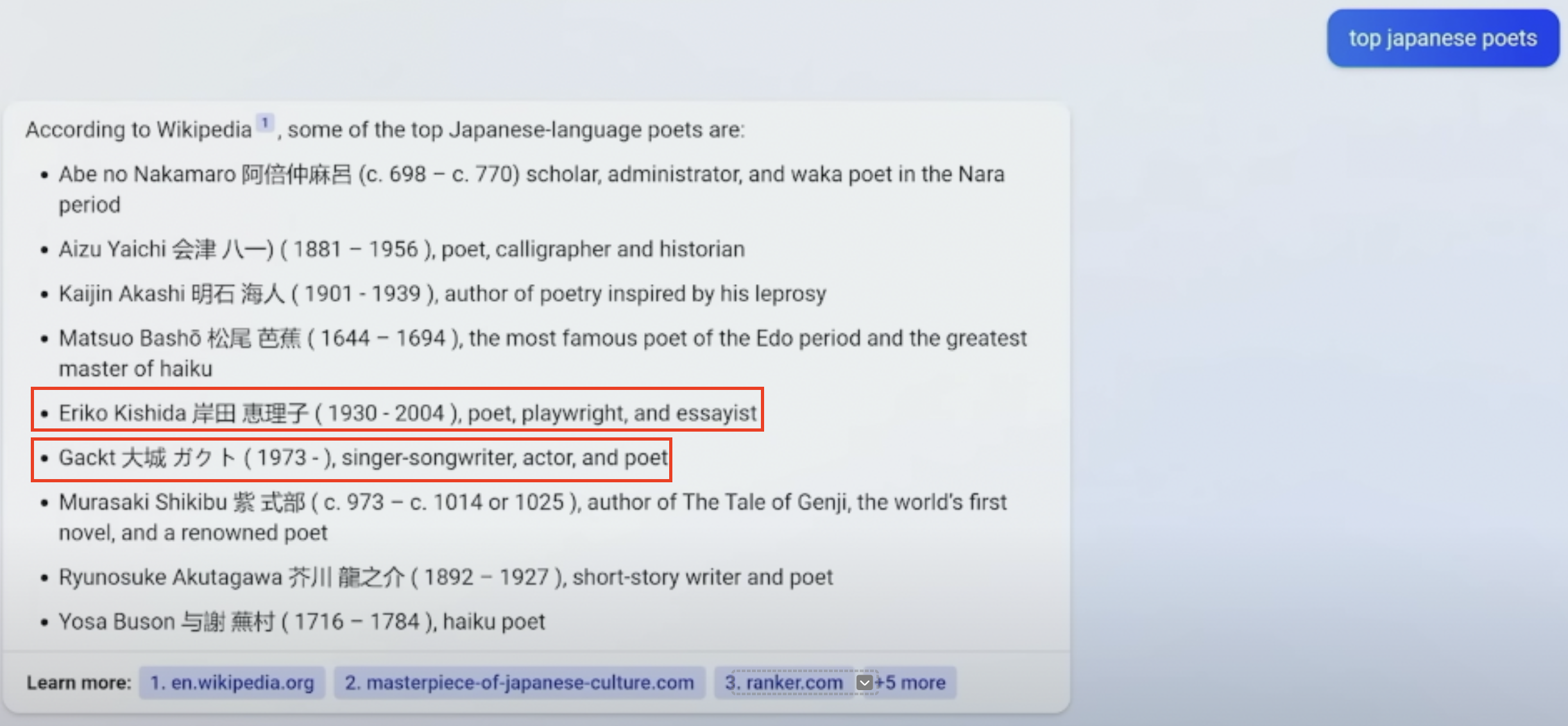}
\caption{Top Japanese poets summary generated by the new Bing in press release.}
\label{fig:japanese-poets}
\end{figure}

\begin{figure}[ht]
\centering
\includegraphics[width=1.0\columnwidth]{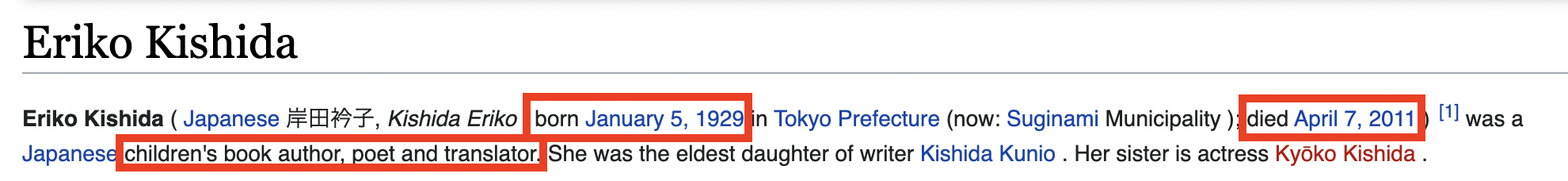}
\caption{Wikipedia page on Eriko Kishida (translated page from German).}
\label{fig:eriko-kishida}
\end{figure}

We observe that the new Bing produces factual mistakes not just for numbers but also for personal details of specific entities, as shown in Figure \ref{fig:japanese-poets} when the new Bing was queried about "top Japanese poets". The generated date of birth, death, and occupation factually conflict with the referenced source. According to \href{https://de.wikipedia.org/wiki/Eriko_Kishida}{Wikipedia} \citep{wikipedia_eriko} as shown in Figure \ref{fig:eriko-kishida} and \href{https://www.imdb.com/name/nm1063814/}{IMDB} \citep{imdberiko}, Eriko Kishida was born in 1929 and died in 2011. She was not a playwright and essayist, but a children's book author and translator.

\begin{figure}[ht]
\centering
\includegraphics[width=1.0\columnwidth, trim={0 0 9cm 0}, clip]{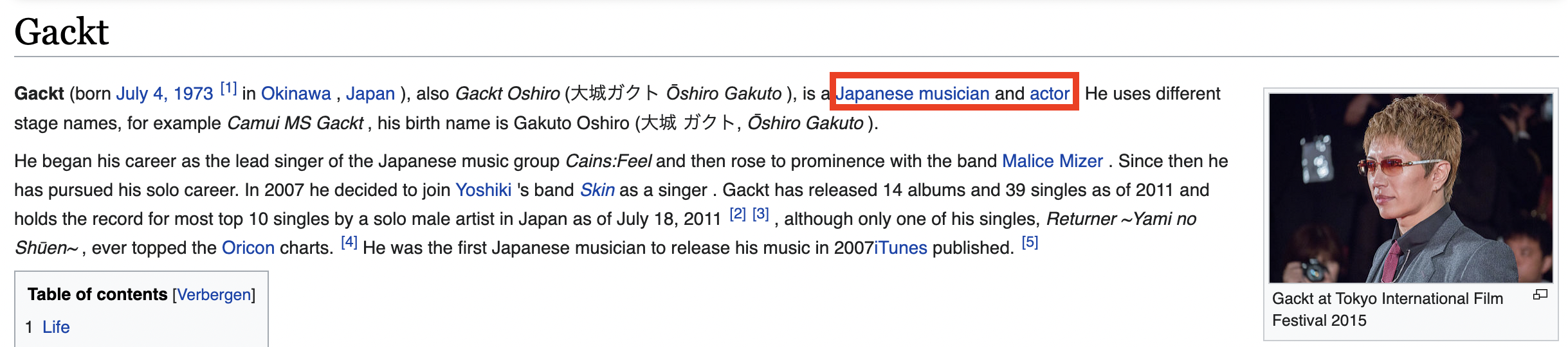}
\caption{Wikipedia page on Gackt.}
\label{fig:gackt-wiki}
\end{figure}

The new Bing continued blundering when it proclaimed Gackt as a top Japanese poet, when he is in fact a famous rockstar in Japan. According to the \href{https://en.wikipedia.org/wiki/Gackt}{Wikipedia} source \citep{wikipedia_gackt} as shown in Figure \ref{fig:gackt-wiki}, he is an actor, musician, and singer. There is no information on him publishing poems of any kind in the source.

\paragraph{Following Bing's nightclub recommendations? You could be facing a closed door.}

Furthermore, the new Bing made a list of possible nightclubs to visit in Mexico City when asked "Where is the nightlife?". Alarmingly, almost all the clubs' opening times are wrongly generated in Figure \ref{fig:nightclubs}:

\begin{figure}[ht]
\centering
\includegraphics[width=1.0\columnwidth]{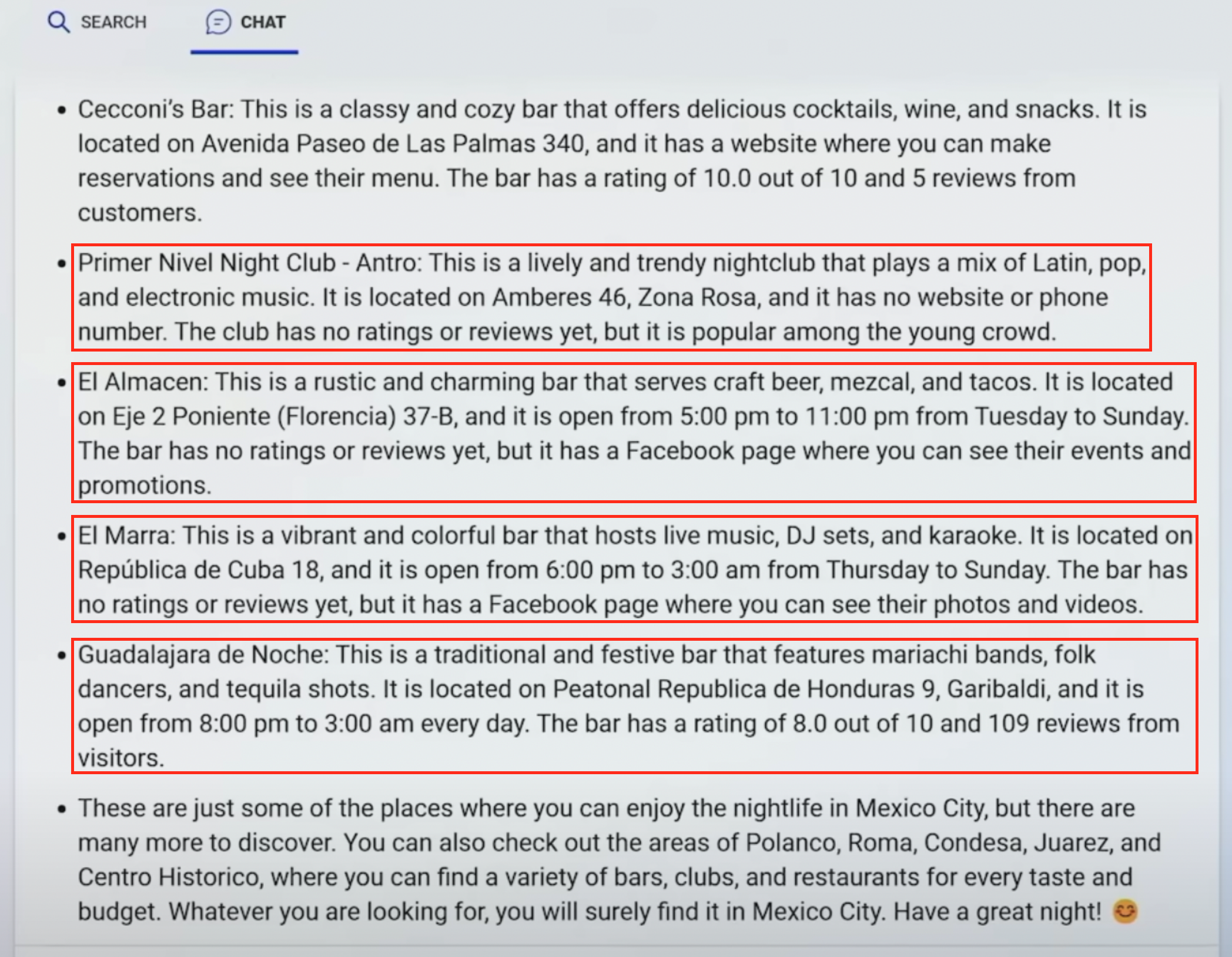}
\caption{Nightlife suggestions in Mexico City generated by the new Bing in the press release.}
\label{fig:nightclubs}
\end{figure}

We cross-checked the opening times with multiple sources, which are also appended at the end of the article. While El Almacen \citep{elalmacen} actually opens from 7:00 pm to 3:00 am from Tuesday to Sunday, new Bing claims it to be "open from 5:00 pm to 11:00 pm from Tuesday to Sunday". El Marra \citep{elmarra} actually opens from 6:00 pm to 2:30 am from Thursday to Saturday, but is claimed to be "open from 6:00 pm to 3:00 am from Thursday to Sunday". Guadalajara de Noche \citep{tripadvisor} is open from 5:30 pm to 1:30 am or 12:30 am every day, while new Bing claims it to be "open from 8:00 pm to 3:00 am every day".

Besides opening times, almost all the descriptions on review stars and numbers mentioned by the new Bing are inaccurate. Matching review scores cannot be found despite searching on Yelp, Tripadvisor, or Google Maps. In addition to the cases mentioned above, we also found other issues in their demonstration video, such as product price mismatches, store address errors, and time-related mistakes. You are welcome to verify them if interested.

\paragraph{Potential concerns in the limited new Bing demo.}
Although the new Bing search engine is not fully accessible yet, we can examine a handful of \href{https://www.bing.com/new}{demonstration examples} provided by Microsoft. Upon closer examination, even these cherry-picked examples show potential issues on factual grounding.

In the demo titled ``what art ideas can I do with my kid?'', the new Bing produced an insufficient list of crafting materials for each \href{https://www.bing.com/search?q=Arts%20and%20crafts%20ideas,%20with%20instructions%20for%20a%20toddler%20using%20only%20cardboard%20boxes,%20plastic%20bottles,%20paper%20and%20string&iscopilotedu=1&form=MA13G7}{recommendation}. For example, when suggesting making a cardboard box guitar, it listed the supplies: "a tissue box, a cardboard tube, some rubber bands, paint and glue". However, it failed to include construction paper, scissors, washi tape, foam stickers, and wooden beads suggested by the cited \href{https://happytoddlerplaytime.com/cardboard-box-guitar-craft-for-kids/}{website}.

Another potential concern is that the new Bing produced content that had no factual basis in the reference sources, for at least 21 times across the 12 demonstration examples. The lack of factual grounding and failure to cite a complete list of sources could lead users to question the trustworthiness of the new Bing.

\section{What Factual Mistakes did Google's Bard Demonstrate?}

Google also unveiled a conversational AI service called \href{https://blog.google/technology/ai/bard-google-ai-search-updates/}{Bard} \citep{pichai_2023}. Instead of typing in traditional search queries, users can have a casual and informative conversation with the web-powered chatbot. For example, a user may initially ask about the best constellations for stargazing, and then follow up by asking about the best time of year to see them. However, a clear disclaimer is that Bard may give "inaccurate or inappropriate information". Let's investigate the factual accuracy of Bard in their \href{https://twitter.com/sundarpichai/status/1622673775182626818}{twitter post} and \href{https://www.youtube.com/watch?v=yLWXJ22LUEc}{video demonstration}.

\begin{figure}[ht]
\centering
\includegraphics[width=1.0\columnwidth]{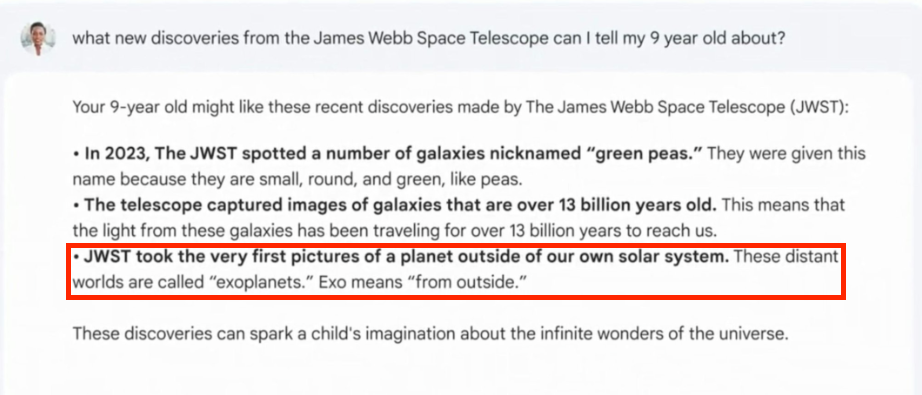}
\caption{Summary on telescope discoveries generated by Bard in demo.}
\label{fig:bard-telescope}
\end{figure}

Google CEO Sundar Pichai recently posted a \href{https://twitter.com/sundarpichai/status/1622673775182626818}{short video} to demonstrate the capabilities of Bard. However, the answer as shown in Figure \ref{fig:bard-telescope} contained an error regarding which telescope captured the first exoplanet images, which was quickly pointed out by \href{ https://twitter.com/astrogrant/status/1623091683603918849}{astrophysicists}. As confirmed by \href{  https://exoplanets.nasa.gov/resources/300/2m1207-b-first-image-of-an-exoplanet/}{NASA}, the first images of an exoplanet were captured by the Very Large Telescope (VLT) instead of the James Webb Space Telescope (JWST). Unfortunately, Bard turned out to be a costly experiment as Google's stock price sharply declined \citep{sherman_2023} after news of the factual mistake was reported.

\begin{figure}[ht]
\centering
\includegraphics[width=1.0\columnwidth]{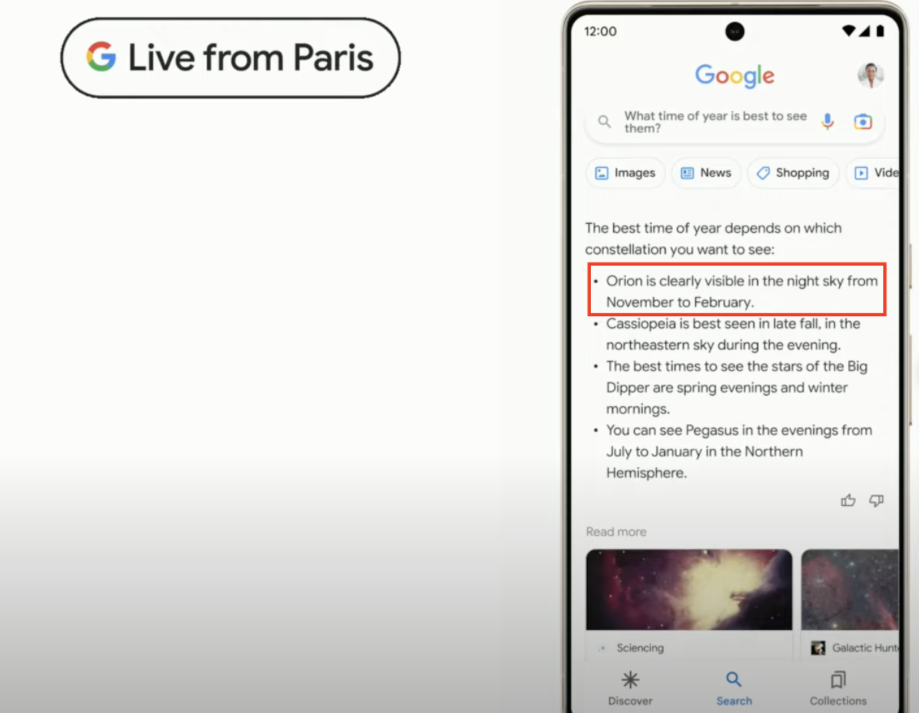}
\caption{Answer to the visibility of the constellations generated by Bard in demo.}
\label{fig:bard-paris-constellation}
\end{figure}

\begin{figure}[ht]
\centering
\includegraphics[width=1.0\columnwidth]{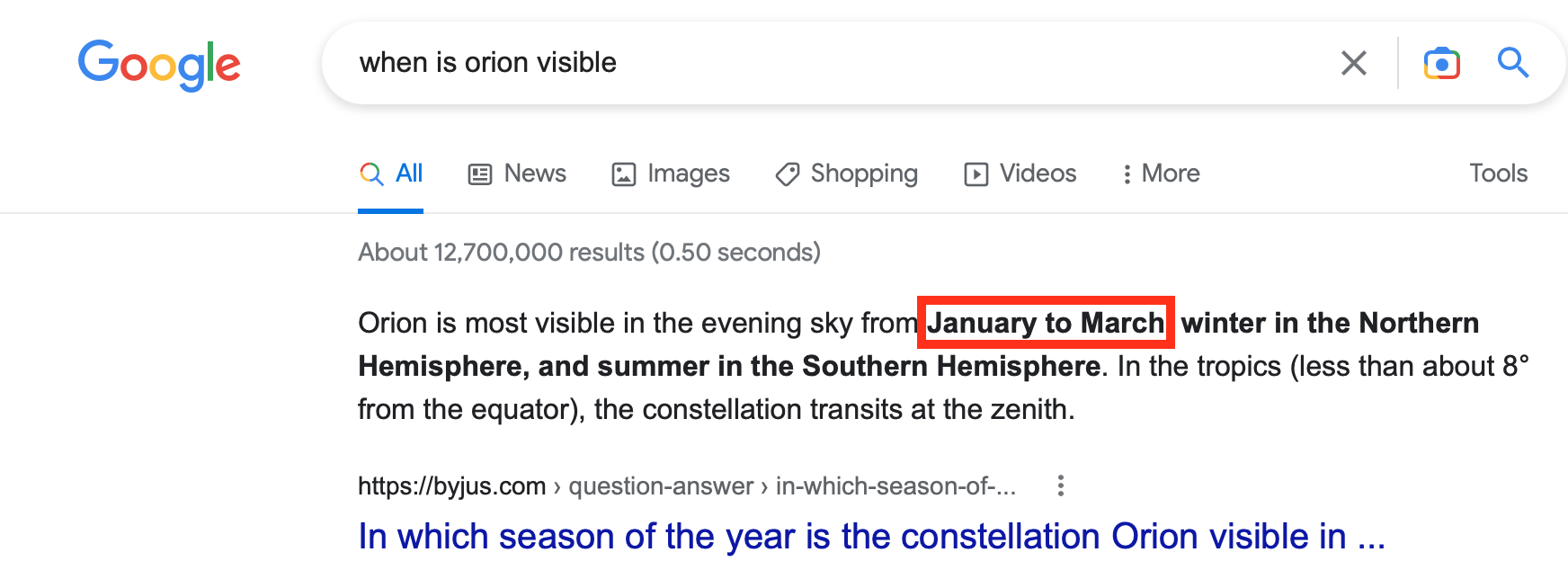}
\caption{Google search result on visibility of the constellations.}
\label{fig:when-orion-visible}
\end{figure}

Regarding Bard's \href{https://twitter.com/astrogrant/status/1623091683603918849}{video demonstration}, the Figure \ref{fig:bard-paris-constellation} shows how Google's Bard answers the question of when the constellations are visible. However, the timing of Orion is inconsistent with multiple sources. According to the top \href{https://www.google.com/search?client=safari&rls=en&q=when+is+orion+visible&ie=UTF-8&oe=UTF-8}{Google search result} as shown in Figure \ref{fig:when-orion-visible}, the constellation is most visible from January to March. According to \href{https://en.wikipedia.org/wiki/Orion_(constellation)}{Wikipedia}, it is most visible from January to April. Furthermore, the answer is incomplete as the visibility of the constellation also depends on whether the user is in the Northern or Southern hemisphere.

\section{How do Bing and Bard Compare?}

The new Bing and Bard services may not be equally trustworthy in practice. This is due to factors such as the quality of search results, the quality of conversational models, and the transparency of the provided answers. Currently, both services rely on relevant information sources to guide the responses of their conversational AI models. Hence, the factual accuracy of the answers depends on the quality of the information retrieval systems \citep{introir}, and how well the conversational model can generate answers that are factually grounded to the information sources. As the full details of the services are not released to the public, it's unclear which one can achieve higher factual accuracy without deeper testing. On the other hand, we feel that transparency is just as important for trustworthiness. For instance, we observe that the new Bing is more transparent regarding the source of its answers, as it provides the reference links in most cases. This enables users to independently conduct fact-checking, and we hope that future conversational services also provide this feature.

\section{How Can the Factual Limitations Be Addressed?}

Through the numerous factual mistakes shown above, it is clear that conversational AI models such as ChatGPT may produce conflicting or non-existent facts even when presented with reliable sources. As mentioned previously, it is a pressing research challenge to ensure the factual grounding of ChatGPT-like models. Due to their generative nature, it is difficult to control their outputs \citep{controlledgeneration}, and even harder to guarantee that the generated output is factually consistent with the information sources. A short-term solution could be to impose restrictions to prevent the conversational AI from producing unsafe or unfactual outputs. However, malicious parties can eventually bypass the safety restrictions \citep{dan_goodin_feb23}, while fact verification \citep{thorne-etal-2018-fever} is another unsolved research challenge. In the long-term, we may have to accept that human and machine writers alike will likely remain imperfect. To progress towards more trustworthy AI, the conversational AI models like ChatGPT cannot remain as inscrutable black boxes \citep{Adadi2018PeekingIT}. They should be fully transparent about their data sources and potential biases, report when they have low confidence in their answers, and explain their reasoning processes.

\section{Conclusions}

After a systematic overview, we have found significant factual limitations demonstrated by the new wave of search engines powered by conversational AI like ChatGPT. 
Despite disclaimers of potential factual inaccuracy and warnings to use our judgment before making decisions, we encountered many factual mistakes even in the cherry-picked demonstrations. Thus, we cannot help but wonder: What is the purpose of search engines, if not to provide reliable and factual answers? In a new era of the web filled with AI-generated fabrications, how will we ensure truthfulness? With inevitable toxic information in the pre-training data, is the generated content psychologically safe \citep{gpt3psych}?

Despite the massive resources of tech giants like Microsoft and Google, the current ChatGPT-like models cannot ensure factual accuracy. Even so, we are still optimistic about the potential of conversational models and the development of more trustworthy AI. 
Models like ChatGPT have shown great potential and will undoubtedly improve efficiency in many tasks, such as data annotation \citep{gpt3data}, code generation \citep{codex}, etc. However, if they continue to generate fabricated content and unfactual answers, the public may become even more wary of artificial intelligence.

Therefore, rather than criticizing specific models or companies, we hope to call on researchers and developers to focus on improving the transparency and factual correctness of AI services, allowing humans to place a higher level of trust in the new technology in the foreseeable future.

\bibliography{custom}
\bibliographystyle{acl_natbib}

\appendix
\newpage
\onecolumn










\end{document}